\documentclass[11pt,a4paper]{article}
\usepackage[hyperref]{emnlp2020}
\usepackage{times}
\usepackage{graphicx}
\usepackage{amsmath}
\usepackage{graphics}
\usepackage{latexsym}
\usepackage{enumitem}

\usepackage{tabularx}
\usepackage{microtype}
\aclfinalcopy
\def\vs#1{\vspace{0pt}}

\def\aclpaperid{942} %  Enter the acl Paper ID here

\title{Incorporating Stylistic Lexical Preferences in Generative\\ Language Models}

\author{Hrituraj Singh \\
  Adobe Research\\
  \texttt{hrisingh@adobe.com} \\\And
  Gaurav Verma \\
  Adobe Research  \\
  \texttt{gaverma@adobe.com} \\\And
  Balaji Vasan Srinivasan \\
  Adobe Research\\
  \texttt{balsrini@adobe.com}\\
  }

\date{30th Novemeber, 2019}

\begin{document}
\maketitle
\begin{abstract}\vs\vs
While recent advances in language modeling has resulted in powerful generation models, their generation \textit{style} remains implicitly dependent on the training data and can not emulate a specific target style. Leveraging the generative capabilities of a transformer-based language models, we present an approach to induce certain target-author attributes by incorporating continuous multi-dimensional lexical preferences of an author into generative language models. We introduce rewarding strategies in a reinforcement learning framework that encourages the use of words across multiple categorical dimensions, to varying extents. Our experiments demonstrate that the proposed approach can generate  text that distinctively aligns with a given target author's lexical style. We conduct quantitative and qualitative comparisons with competitive and relevant baselines to illustrate the benefits of the proposed approach. 

% Given the recent advances in the generation capabilities of Transformer-based models and the use of reinforcement learning in generating tuned text, we present a reinforcement learning based framework to fine-tune a pre-trained Transformer-based language model on author specific characteristics to adapt the language model's generation to target author's lexical style. Our approach leverages the ability of reinforcement learning based text generation to \textit{not} rely on a huge target text corpora or a differentiable objective function. While there has been some success with employing RL in RNN based language models (LM), our approach is one of the first to demonstrate similar progress on Transformer-based LMs. Furthermore, the current approaches are limited in their style transfer ability with most works transferring sentiment (positive/negative) or one-dimensional feature (e.g. formality/informality) in sentence. Our approach aims to capture the target author's  multi-dimensional continuous lexical style. \GV{TODO}
\end{abstract}

\section{Introduction}\vs\vs\vs\vs
%There has been an increased interest in controlling various aspects of text in generation tasks, e.g. sentiment \cite{li2018delete}, formality \cite{rao2018dear}, style \cite{shen2017style}. 
With recent advances in unconstrained language generation \cite{gpt, gpt2, gpt3}, an emerging direction is to adapt such pre-trained language models to follow certain stylistic constraints \cite{wang2019harnessing, syed2019adapting}. These approaches rely on the inherent properties of the training corpus to tailor generation to target characteristics; for example, implicitly learning author-stylized text generation by training on author-specific corpus \cite{syed2019adapting} and learning to generate formal text \cite{wang2019harnessing}. However, it is desirable to have explicit control over certain stylistic aspects in such generation, for e.g., emulating lexical choices of an author in a generation, capturing syntactic constructs, inducing sentential preferences (active vs. passive) in generation. To this end, we propose an approach to adapt a pre-trained Transformer-based language model \cite{vaswani2017attention}, specifically GPT-2 \cite{gpt2}, to generate text that aligns with given lexical elements of style by providing \textit{explicit} rewards in a reinforcement learning framework. 
 
Reinforcement learning (RL) has been successfully applied to several natural language generation tasks like summarization \cite{paulus2017deep} and paraphrase generation \cite{li2018paraphrase}. RL overcomes the `exposure bias' \cite{ranzato2015sequence} in cross-entropy based training of language models and allows for optimization with respect to non-differentiable objectives. However, existing explorations around the use of RL in generation tasks have been limited to RNN-based models due to issues surrounding stabilization of RL training on Transformer models \cite{parisotto2019stabilizing}. \citeauthor{parisotto2019stabilizing} further conclude that a Transformer requires reordering of the normalization layer from output to the input streams along with a gated mechanism instead of residual connections to stabilize its training. Building on this, we leverage the shifted position of the normalization layers in GPT-2 to train an RL framework with GPT-2 for aligning generated text to target lexical characteristics.

Recent work by \citeauthor{ziegler2019fine} (\citeyear{ziegler2019fine}) explored RL frameworks with GPT-2 to generate text with different styles. However, they treat their target characteristics (sentiment and descriptiveness) as a binary variable (viz. +ve/-ve). It is non-trivial to extend their work to generate lexically-aligned text, since each of the target dimensions is a continuous value. %, posing a unique challenge.
Our task further requires \textit{simultaneously} aligning along multiple lexical dimensions calling for a rewarding strategy that accounts for multiple dimensions. 
%An author's writing style is an amalgamation of variety of elements ranging from narrative to sentential to lexical. Our objective, in this paper, is however constrained to lexical style \textit{only}.
% have explored reward learning and optimization on binary characteristics somewhat concurrently to our work. However, their work is primarily aimed at achieving one-dimensional characteristic in the output with a binary distinction; for example, whether the generated output should have positive sentiment or negative sentiment. On the other hand, our work is focused on achieving multidimensional lexical targets to emulate author's style with a continuous variation in each target characteristic.
% Thus, to the best of our knowledge, our work is one of the first to show results on pretrained transformer model with RL based tuning. 
To this end, our key contributions are:
\noindent$\mathbf{(1)}$ an RL framework that introduces lexical style elements in a Transformer-based language generation model; $\mathbf{(2)}$ a rewarding scheme that incorporates continuous multi-dimensional lexical elements; $\mathbf{(3)}$ extensive experiments on multiple authors to show the efficacy of our approach to align generation to an author's lexical preferences. \vs\vs

\section{Author's Lexical Style}\label{sec:Style}\vs\vs\vs\vs\vs\vs

An author's writing style is a combination of several factors that include, but are not limited to, their lexical preferences, syntactic and sentential choices (e.g., active vs. passive voice, use of detached adjectival clause), discourse structure, narrative style, etc. To perfectly reproduce a given author's style, the language generation model should operate in accordance to all these factors. However, we limit ourselves to replicating an author's \textit{lexical} style, which refers to their writing choices at \textit{word-level}. 

\citeauthor{lexical} (\citeyear{, brooke2013hybrid, lexical}) enumerate lexical style elements into \textbf{subjective, objective, literary, colloquial, abstract and concrete} categories. An author's choices of words in these categories define their lexical style. For example, Rudyard Kipling, known for classics of children’s literature, had a higher tendency to use more concrete words (like, gongs, rockets, torch) unlike Abraham Lincoln,
who being a political writer, used more abstract words (like freedom, patriotism) \cite{verma2019lexical, syed2019adapting}. Since an author's style is an amalgam of preferences along these dimensions, our goal is to ensure simultaneous alignment to these multi-dimensional lexical preferences of an author. 

To quantify a target author's lexical preferences, following
\citeauthor{lexical} (\citeyear{lexical}), we compute normalized pointwise mutual information index (PMI) of each vocabulary word with every seed word of $6$ categories using their co-occurrences in the Emobank \cite{emobank} corpus yielding a \textit{raw style score} for each category, for each word in the vocabulary. An author's affinity to a particular style is calculated by the fraction of positive style words in their corpus, yielding a $6-$dimensional vector with each value $\in [0, 1]$. \vs\vs\vs\vs
% to ensure alignment along a target author's lexical vector. 
% we utilize the list of seed words for the six categories above and compute normalized pointwise mutual information index (PMI) by utilizing the co-occurrence matrix created using EmoBank corpus\cite{emobank} to obtain the style score for each of the categories above thus forming a six dimensional vector $L$ representing the style of each word.
 Unlike \cite{lexical}, we do not consider formality-informality pair due to the imbalance in the number of seed lexicons provided for formality and informality. With a suitable seed lexicon, our approach however can be extended to these two characteristics as well.

\section{Proposed Approach}\label{sec:approach}\vs\vs\vs\vs
A language model (LM) $G$ models generation of a sequence $X$ as a task of sampling tokens $x_0$ to $x_m$. Here, a token $x_i$ is sampled from a probability distribution conditioned on the previously generated tokens $x_0$ to $x_{i-1}$,
\begin{equation}
    G(X \mid C) = \prod_{i=0}^{m} G(x_i \mid x_0..x_{i-1}, C)
\end{equation}
where, $C=c_0,\ldots,c_n$ is the context for generation given by the input prompt to $G$ which provides a sense of broader restriction on generation of $X$.\\

\noindent\textbf{Episode Unrolling: }
An agent ($G$ in our case) in an RL framework, learns a \textit{policy} $\pi$ to perform a set of \textit{actions} $a_i$ (i.e. generating tokens) resulting in a change of its \textit{states}. The policy's action $a_i$ at a state $S_{i-1} : \{C,x_0, \ldots, x_{i-1}\}$ results in the generation of a token $x_i$ which takes the model to state $S_i : \{C,x_0 \ldots x_{i}\}$. We refer to the sequence of tokens $E: \{a_0, a_1...,a_t\}$ generated by the LM as it arrives at the terminal state $S_t$ as an \textit{episode}. Instead of relying on a linguistic terminal property (such as end of sentence), we utilize length of the generated sequence as the terminal property of a state. This ensures that the lexical statistics across episodes are consistent while computing the rewards.

\begin{figure}[t]
\centering\vs\vs\vs\vs
\includegraphics[width=\linewidth]{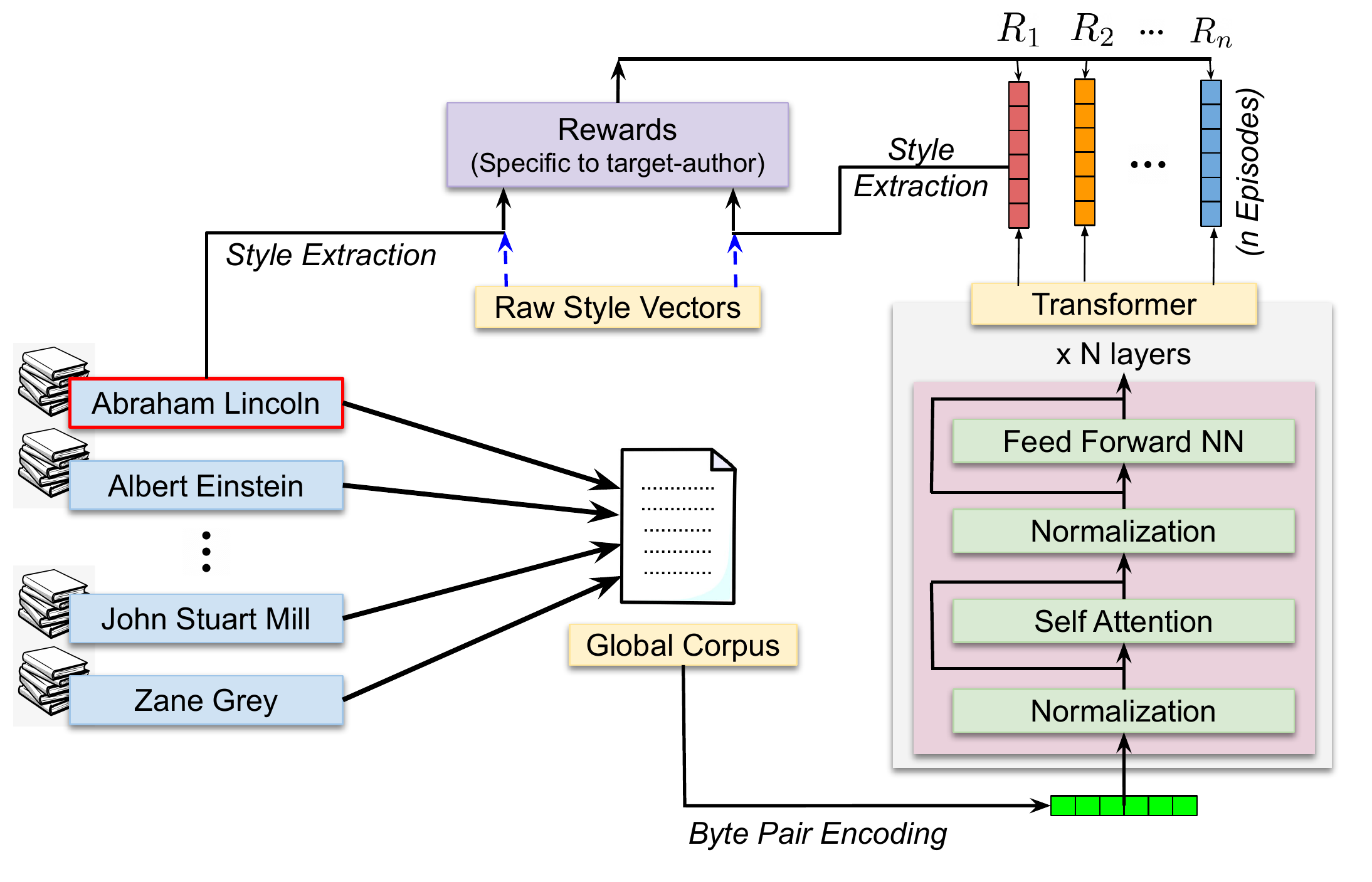}\vs\vs\vs\vs\vs\vs 
\label{proposed}
\caption{\small{\textit{\textbf{Overview of the proposed approach. We aggregate a global corpus using a fixed number of paragraphs from each author. 
The policy generates $n$ episodes for each input in the corpus. Each episode is rewarded with $R_i$ based on its deviation from the target author's lexical style.}}}}
\end{figure}

For each context $C$, we unroll $N$ episodes, $X_1$ to $X_N$, enabling the policy to \textit{explore} the space better. Unlike the traditional multinomial sampling, we use the \textit{nucleus sampling} \cite{nuclear_sampling} that restricts the sampling to the `nucleus' of the distribution for generating episodes. By dissuading the choices from the long tail of the distribution, \textit{nucleus sampling} allows the framework to exploit the policy's learning (so far) and hit a balance between \textit{exploration} and \textit{exploitation}.\\

\noindent\textbf{Rewarding Strategy: }
%We employ an RL-based rewarding strategy to realize our non-differentiable objective of incorporating lexical characteristics of an author. 
%For a given context $C$, and an episode $X \sim \pi (. \mid C)$ generated with $C$, we define a reward $R(C, X)$. % with an objective to maximize the expected reward,
%${E}[R] = {E}_{C, X \sim \pi(.\mid C)}[R(C,X)].$
\citeauthor{lexical} (\citeyear{lexical}) quantify lexical characteristics using paragraph-level statistics. Extending this, we reward the model with $r$ at the final action $a_t$ of the episode and give $0$ reward to the intermediate actions $a_i, i\neq t$ -- where $t$ is the terminal step. The overall reward $R_i$ for an \textit{action} at a time step \textit{i}, where the terminal time step is \textit{t} and immediate reward received at step $i$ is $r_i$, is given by discounting the future rewards \cite{sutton2018reinforcement} with a factor $\gamma$, 
\begin{equation}
    R_i = r_i + \gamma*r_{i+1} + \gamma^2*r_{i+2} ....  + \gamma^{t-i}*r_t
\end{equation}
% \GV{TODO-check}
Setting $\gamma = 1$, we distribute the reward uniformly over the entire sequence; all the actions in a particular sequence receive same award irrespective of their position. Since the style is considered at the sequence level, the position of token is irrelevant as long as fluency is maintained.\\

\noindent\textbf{Defining Rewards:}
We define the \textit{inclination} of a token ($L_{incl}^i$) to a target style category to be $1$ if its raw style score (from Section \ref{sec:Style}) is positive and $0$ otherwise. Averaging these across all the tokens in an episode yields the lexical alignment score $L_{epi} = \frac{1}{m} \sum_{i=0}^{m}L_{incl}^i$. Since an author's lexical style is an amalgam of several characteristics, we use a Root Mean Squared error against the author's statistics as our aggregated reward for all $6$ elements and enabling a continuous adjustment of generation across the target elements. Given the lexical statistic $L_{tar}$ of a target author, the reward $r = \frac{1}{rmse + \epsilon}$,where,  $rmse = \frac{\|L_{epi} - L_{target}\|}{\sqrt{6}}$ and $\epsilon$ is a factor used to avoid division by zero.

We use our rewards in a modified \textbf{self-critical sequence training }setup \cite{scst} because this was the most stable framework in our exploration. In our experiments, we have described the other frameworks we explored along with our intuitions on their failures in our problem. A multi-dimensional tuning requires more deviation from an existing policy compared to tuning a single dimension, calling for more exploration, enabled by our episode unrolling. For a context $C$, the mean reward from $N$ unrolled episodes is used as the baseline reward $r_b$ to reduce the variance during training. Following REINFORCE \cite{reinforce}, we minimize the following loss function ,
\begin{equation}
    \small
    J(\theta) = - (r-r_b)\sum_{i=0}^{m} log(\pi_{\theta}(x_i \mid x_0, \ldots, x_{i-1}, C))
\end{equation}
where $\theta$ are policy parameters. We scale the rewards for a given context to zero-mean and unit-variance across the $N$ episodes. %, keeping it \textit{approximately} in $[-1, 1]$. 
Following \citeauthor{ranzato2015sequence} (\citeyear{ranzato2015sequence}), we minimize cross entropy on the tokens from $C$ along with $J(\theta)$ every $5$ contexts (\textit{i.e.} $5N$ episodes), so that the model does not deviate and retains its fluency. During this step, our loss is a weighted sum of cross entropy loss and RL loss (empirically set to $0.5$ and $1.0$). \vs\vs

\begin{table*}[bth]
\centering
\vs\vs
\scalebox{0.62}{%
\begin{tabular}{l|ll|ll|ll|ll|ll|ll|ll|c}
 \textbf{Model} & \multicolumn{2}{c}{\textbf{Literary}} &\multicolumn{2}{c}{\textbf{Colloquial}}& \multicolumn{2}{c}{\textbf{Abstract}}& \multicolumn{2}{c}{\textbf{Subjective}}& \multicolumn{2}{c}{\textbf{Concrete}} &  \multicolumn{2}{c}{\textbf{Objective}}&\multicolumn{2}{c}{\textbf{Overall}}&\textbf{Perplexity} \\ \hline
 & \textit{abs}& \textit{rel}& \textit{abs}& \textit{rel}& \textit{abs}& \textit{rel}& \textit{abs}& \textit{rel}& \textit{abs}& \textit{rel}& \textit{abs}& \textit{rel}& \textit{abs}& \textit{rel}\\ \hline%\hdashline
\textbf{GPT-2 (\citeauthor{gpt2})} & 0.246&1.117 &0.081& 1.880 &0.127&1.708 &0.323& 0.430& 0.270&1.778& 0.064&2.413&0.297&0.518& 53.82 \\ %\hdashline
\textbf{GPT-2 (Baseline)} & 0.236 &1.255&\textbf{0.060}&1.761&\textbf{0.101}&1.696& 0.293 &0.579&0.251&1.700& 0.088&2.340&0.283&0.518&\textbf{37.54 } \\
\textbf{GPT-2 + FT} & 0.242& 1.192 &0.070& 1.757& 0.115&1.622 &0.298&0.467 &0.237&1.700 &0.066& 2.321&0.283&0.503 &38.43 \\
\textbf{GPT-2 + RL (5K Episodes)} &0.174&0.906 &0.062&1.395& 0.120&1.211 &0.222& 0.323&0.147& 1.295&0.047&1.572&0.221&0.372&38.81\\
\textbf{GPT-2 + RL (10K Episodes)} &\textbf{0.160}&0.869& 0.066&\textbf{1.328}& 0.118&\textbf{1.169}& \textbf{0.212}&0.323& 0.143&1.253& 0.046&1.525&\textbf{0.213}&0.359&38.57 \\
\textbf{GPT-2 + FT + RL (5K Episodes)} & 0.176& 0.922 &0.067& 1.433 &0.129&1.231 &0.232& 0.347&0.154&1.296 &\textbf{0.041}&1.649 &0.226&0.382 &38.91 \\
\textbf{GPT-2 + FT + RL (10K Episodes)}&0.162& \textbf{0.869}&0.064&1.344 &0.126&1.170 &0.213&\textbf{0.314} &\textbf{0.139}& \textbf{1.245}&0.048& \textbf{1.503}&0.214&\textbf{0.358} &38.59 \\
\hline
\end{tabular}}
\vs\vs\vs\vs\vs\vs\caption{\label{result-table} \small{Results from our Quantitive Evaluation}: `\textit{abs}' error for each dimension is calculated as the absolute difference between the target value and obtained value for that dimension while `\textit{rel}' error is the absolute deviation of the relative order of the dimension based on the $L_1$ norm. `Overall' \textit{abs} is the RMSE between output and target 6 dimensional vectors while `Overall' \textit{rel} is the average \textit{rel} error across all dimensions. Perplexity indicates the deviation of the model from its fluent generation.\vs\vs\vs\vs\vs\vs}
\end{table*}

\vs\vs\vs\vs
\section{Experiments} 
\vs\vs\vs
We used the $2,857$ books of $142$ authors in Gutenberg corpus \cite{guteneberg} and divided each author's corpus into \textit{train} and \textit{test} sets. We concatenate $50$ paragraphs from each author's train corpus and use this for fine-tuning with language modelling loss for each author. To evaluate our model on unseen data, we set aside a subset of $5$ paragraphs from each author's test corpus to be used as test context. Having contexts from all authors removes any bias from author-specific contexts. 

We compute the average lexical vectors $L_{avg}$ for all authors and retain top $10$ authors with maximum deviation from $L_{avg}$ 
for our experiments: \textit{Charles Darwin, Albert Einstein, Michael Faraday, John Maynard Keynes, Abraham Lincoln, John Locke, John Stuart Mill, Beatrix Potter, Bertrand Russell, and Herbert Spencer}.
We use the $117M$ parameters version of the GPT-2 \cite{gpt2} trained on WebText corpus and $50,257$ token invertible byte pair encoding to preserve capitalization and punctuation \cite{bpe}. The model is a $12-$layers $12-$head Transformer with embedding size of $768$. Finetuning\footnote{Trained on 8 V100 GPUs for 12 hours} GPT-2 on entire Gutenberg corpus yields \textbf{GPT-2 (Baseline)}. Finetuning further for one epoch on the target author's corpus with Causal Language Modelling (CLM) loss yields \textbf{GPT-2 + FT}. For RL finetuning, we use a batch size $1$, $10$ episodes for each context, context length $200$, episode length $100$ and $0.05$ as $\epsilon$. 

\begin{figure}[t]
\centering\vs\vs\vs\vs\vs\vs
\includegraphics[width=\linewidth]{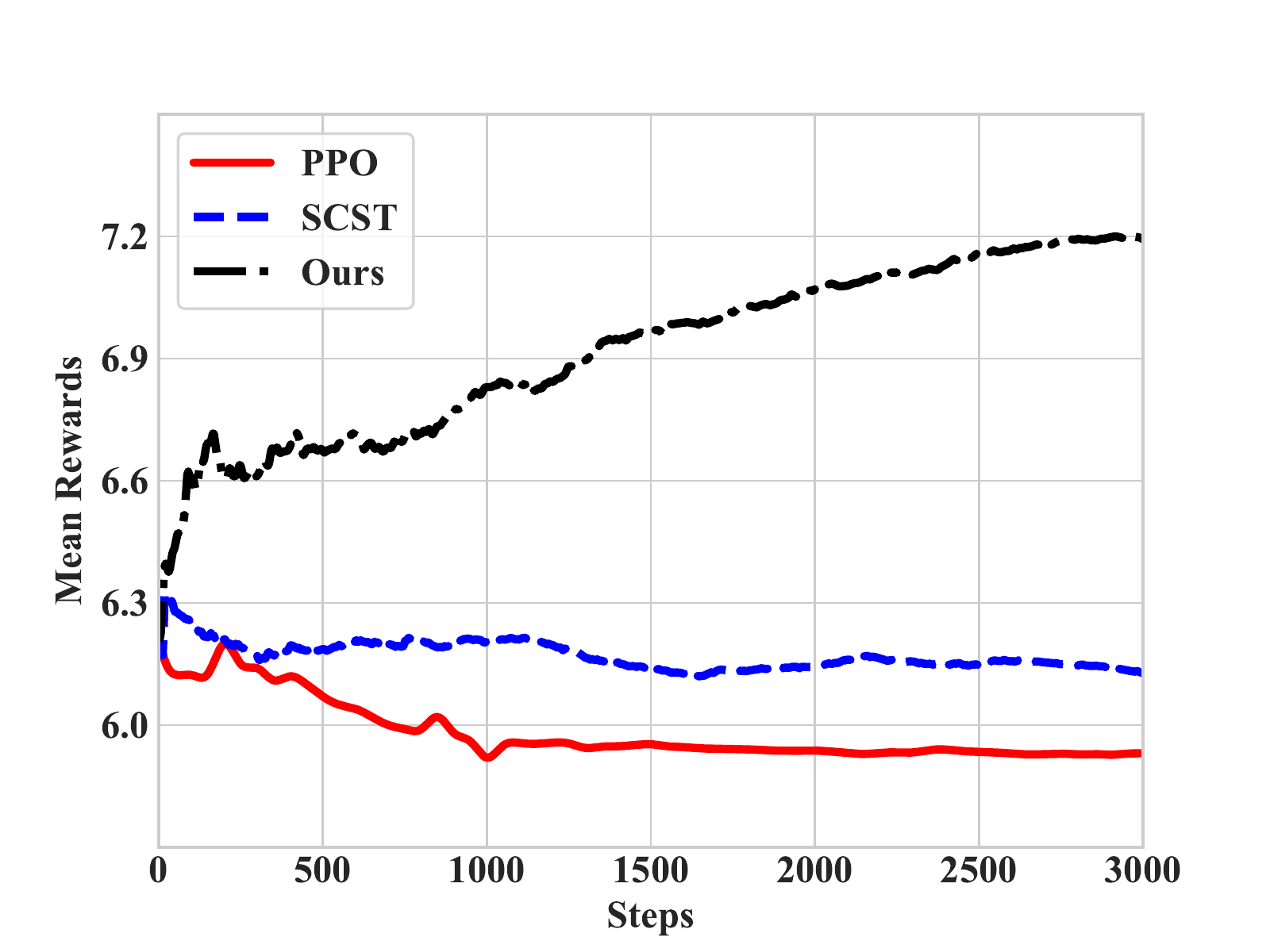}\vs\vs\vs\vs\vs\vs\vs\vs\vs\vs\vs\vs
\caption{\textbf{\textit{Mean Reward Curves as averaged over all ten different authors for the different algorithms}}\label{fig:RewardsProposed}\vspace{-5pt}}
\end{figure}

We explored Self-Critical Sequence Training (SCST) \cite{scst} and Proximal Policy Optimization (PPO) \cite{schulman2017proximal} for our RL setup and chose our episode unrolled SCST because of its stability. Figure \ref{fig:RewardsProposed} shows the mean reward curves averaged over 3 randomly drawn authors; SCST does not help in improvement of rewards, perhaps due to the lesser exploration carried out in the vanilla setup leading to little-to-no-improvement in the lexical style of the generation. PPO rewards goes down a bit and stays almost constant, perhaps due to the failure of critic. Approximating continuous lexical score is challenging unlike binary positive or negative style – the ones for which PPO has been successful \cite{ziegler2019fine}, hence the critic finds it difficult to approximate the value functions. Our episode unrolling based SCST explores enough to improve rewards quickly while frequent cross-entropy objective training ensures that the improvements do not come at cost of fluency. Note that the rewards saturate after a few steps as the nucleus of the distribution gets shifted towards target lexical style.

We calculate the lexical vector $L_{seq}$ for each generated \textit{paragraph} and compute the error against the target author's lexical vector $L_{tar}$. For dimension-specific error, we take absolute difference between target and generated value for each author and average across all $10$ authors. For overall error, we calculate the RMSE between $L_{tar}$ and $L_{seq}$. We report the perplexity of the model on the contexts in the test set to measure its deviation from its general generation capabilities. We also quantify the alignment in relative ordering of target dimensions using the $L_1$ norm between the relative positions of the attributes in $L_{seq}$ and $L_{tar}$. This evaluates the generation of models on their ability to achieve the target author's relative ordering. In Table \ref{result-table} along with our absolute error differences along with the $L_1$ norm as the deviation in the relative order. The overall deviation is computed by adding the $L_1$ norms across all dimensions and dividing by the maximum possible deviation (which is $18$ for a $6$ dimensional vector). Our evaluations show the success of our approach in aligning the generation across all lexical dimensions, while CLM fine-tuning does not yield significant improvements over the baseline. We also notice that our model achieves lexical alignment after going through training on just $5k$ episodes evidenced by an insignificant decrease in error for $10k$ episodes. Our approach also ensures that the model has not lost its general generation capabilities which is evident with a marginal drop in perplexity, indicating a minimal trade-off between the lexical alignment and fluency. Infact, the perplexity scores in our case are very significantly lower than the score obtained with an out of domain GPT-2.

\begin{table*}[t!]
\centering
\small
\begin{tabularx}{1\linewidth}{X}\\
\hline\\
\textbf{Context (originally written by D. H. Lawrence)}: till now he had a tight, hot place in his chest, with suppressing the pain, and holding himself in. There seemed no air when he breathed. But he walked almost lightly.
To his orderly he was at first cold and just and indifferent: he did not fuss over trifles. So that his servant knew practically nothing about him, except just what orders he would give, and how he wanted them obeyed. That was quite simple. Then the change gradually came.
He had served the Captain for more than a year, and knew his duty. This he performed easily, as if it were natural to him. The officer and his commands he took for granted, as he took the sun and the rain, and he served as a matter of course. It did not implicate him personally. I have said that the sole effect of my somewhat childish experiment--that of looking down within the tarn had been to deepen the first \\\\
\hline
\\
\textbf{GPT-2 (\cite{gpt2}):} singular truth of the Pilgrimage of the Infant. I asked how I was to give my obedience, and, having been told that I must respond by force with violence, I found I had nothing else, less a price and hunger, but the unimaginable change of love and the uniformity of his love that he manifested, in his actions and his conduct...\\\\
\hline
\\
\textbf{GPT-2 (Baseline):} singular comfort I had felt to-day. The idea of it came back to me, with all its allurement in the old days, over and over again; that it had been a serviceable accomplishment, but a training is apt to all but overwhelm the rest. That, too, was the fact of the plunge, the sudden crisis of this large one. With the active man and the active woman, with the mistress and the husband, with the shop-lenderer and the governess\\\\
\hline
\\
\textbf{Albert Einstein (GPT-2 + FT):} singular feeling in me--or to strengthen the second. I don't exactly know the two; but both of them, even in the early days of my illness, occurred in the true light. I felt sorry, indeed, for my sailor, for his predicament, and for some of the miserable conditions of the household. Perhaps, in so doing, I had really helped myself. It was not at all like a stock with its cartridges. The quartermaster and his company, the Earl of\\\\
\hline
\\
\textbf{Albert Einstein (Ours):} singular \textbf{feature} of the \textbf{observations} of the \textbf{experimental} method of physics. I shall leave the first of these observations (I don't wish to show that the experiments which have been referred to are successful) to the reader who, during a similar period of the subject, considers the relations of natural and experimental \textbf{phenomena} to the general \textbf{principles} of \textbf{inductive} \textbf{theory}.\\\\
\hline
\\
\textbf{Abraham Lincoln (GPT-2+FT):} singular feeling in me--and to restore the deepness of the conviction that I was a woman; but, in a personal way, I had achieved what was possible only by looking down. The emotion still hovered; and I recollect at that instant a certain feeling--I say feeling--as I struck the buckle. The feeling made me feel that I must be rather an offensive person to be of any use or inconvenience to me. The thought had not, of course, touched\\\\
\hline
\\
\textbf{Abraham Lincoln (Ours)}:  singular aspect of this \textit{pleasant} and yet \textit{intimate} \textit{connection} which that early talk had formed with the man I now occupy, and, instead of being \textit{self-contained} and thoroughly historical, it had quite come to be represented by a carefully and strongly expressed recollection, as an aggregation of our \textit{relations} and the signs and symptoms of an \textit{attachment}. Indeed, the meeting with Meade's \textit{sense of belonging}, as an influence on his future, had been rendered infinitely more \textit{interesting} by the clear \textit{knowledge} of a more definite direct relation to myself\\
\hline
\\
\end{tabularx}
\vs\vs\vs\vs\vs\vs\caption{\label{result-table3}\small{ \textbf{Qualitative Example.} Generations by different models for the same input prompt. Our model is able to effectively incorporate target author's vocabulary without compromising much on fluency. Here, \textbf{Ours} refers to \textbf{GPT-2 + FT + RL (10K)}.}}\vs\vs\vs\vs\vs\vs\vs\vs\vs\vs\vs\vs\vs\vs\vs\vs\vs\vs
\end{table*}

In the qualitative example in Table \ref{result-table3}, note that our method has infused scientific phrases (e.g., \textbf{`observations', `theory', `experimental'}) and abstract phrases (e.g., \textit{`self-contained', `sense of belonging', `intimate connection'}) for Albert Einstein and Abraham Lincoln, respectively. There is also topic-level alignment with the target authors – a concomitant of meeting the target author's lexical preferences. Fine-tuning GPT-2 on Gutenberg corpus induces literary words like `\textit{allurement}', `\textit{serviceable}', `\textit{shop-lenderer}' and `\textit{governess}'; perhaps because Gutenberg corpus contains several literary words not encountered in WebText. Fine-tuning on author-specific corpus (GPT-2 + FT) induces generic stylistic changes, but not necessarily along lexical dimensions. 
Our approach incorporates lexical preferences most evidently. 
% and further reinforcing the trends in Table \ref{result-table}. 

\vs\vs\vs\vs\vs\vs\section{Conclusion and Future Work}\vspace{-5pt}
 
We proposed an approach to incorporate lexical choices of a target author in the generations of a Transformer-based LMs. %While human evaluations are difficult \cite{syed2019adapting}, 
Our quantitative and qualitative evaluations illustrate that our proposed method is successful in aligning lexical characteristics of generation with target author. We believe that our work can also lead to \textit{rewriting} of the input content tailored to certain characteristics, if we can design additional rewards to retain content. We have not performed a complete human evaluation due to the high-level of required expertise among the annotators for this task, as pointed by \citeauthor{syed2019adapting} (\citeyear{syed2019adapting}). Designing the feedback mechanism for such a human evaluation is non-trivial and has been left as a part of the future work along with designing rewarding schemes to capture other author-specific characteristics (e.g., syntactic choices, discourse structure). Despite the lack of such evaluation, these results are promising and offer a plausible line of research for replication of an author's style.

\bibliography{anthology,emnlp2020}
\bibliographystyle{acl_natbib}

\end{document}